\documentclass[times, review, 10pt]{elsarticle}

\usepackage{amssymb}
\usepackage{amsmath}
\usepackage{todonotes}
\usepackage{hyperref}
\usepackage{listings}

\usepackage{algorithm}%
\usepackage{algorithmicx}%
\usepackage{algpseudocode}%
\usepackage{arydshln} % for dashed lines

\definecolor{codegreen}{rgb}{0,0.6,0}
\definecolor{codegray}{rgb}{0.5,0.5,0.5}
\definecolor{codepurple}{rgb}{0.58,0,0.82}
\definecolor{backcolour}{rgb}{0.95,0.95,0.92}
\lstdefinestyle{mystyle}{
    backgroundcolor=\color{backcolour},
    commentstyle=\color{codegreen},
    keywordstyle=\color{codepurple},
    numberstyle=\tiny\color{white},
    stringstyle=\color{blue},
    basicstyle=\scriptsize\ttfamily,
    breakatwhitespace=false,
    breaklines=true,
    captionpos=b,
    keepspaces=true,
    numbers=left,
    numbersep=5pt,
    showspaces=false,
    showstringspaces=false,
    showtabs=false,
    tabsize=2,
    language=Python
}
\begin{document}

\begin{frontmatter}

%% Title, authors and addresses

%% use the tnoteref command within \title for footnotes;
%% use the tnotetext command for theassociated footnote;
%% use the fnref command within \author or \affiliation for footnotes;
%% use the fntext command for theassociated footnote;
%% use the corref command within \author for corresponding author footnotes;
%% use the cortext command for theassociated footnote;
%% use the ead command for the email address,
%% and the form \ead[url] for the home page:
%% \title{Title\tnoteref{label1}}
%% \tnotetext[label1]{}
%% \author{Name\corref{cor1}\fnref{label2}}
%% \ead{email address}
%% \ead[url]{home page}
%% \fntext[label2]{}
%% \cortext[cor1]{}
%% \affiliation{organization={},
%%             addressline={},
%%             city={},
%%             postcode={},
%%             state={},
%%             country={}}
%% \fntext[label3]{}

\title{Fast Calibrated Explanations: \\Efficient and Uncertainty-Aware Explanations \\for Machine Learning Models}

%% use optional labels to link authors explicitly to addresses:
%% \author[label1,label2]{}
%% \affiliation[label1]{organization={},
%%             addressline={},
%%             city={},
%%             postcode={},
%%             state={},
%%             country={}}
%%
%% \affiliation[label2]{organization={},
%%             addressline={},
%%             city={},
%%             postcode={},
%%             state={},
%%             country={}}

\author[ju]{Tuwe Löfström} %% Author name
\ead{tuwe.lofstrom@ju.se}
% \author[unibo,lamb]{Fatima Rabia Yapicioglu} %% Author name
% \author[unibo,lamb]{Alessandra Stramiglio} %% Author name
\author[unibo]{Fatima Rabia Yapicioglu} %% Author name
\author[unibo]{Alessandra Stramiglio} %% Author name
\author[ju]{Helena Löfström} %% Author name
\author[unibo]{Fabio Vitali} %% Author name

%% Author affiliation
\affiliation[ju]{organization={Jönköping AI Lab, Department of Computing, Jönköping University},%Department and Organization
            % addressline={}, 
            city={Jönköping},
            % postcode={}, 
            % state={},
            country={Sweden}}            
\affiliation[unibo]{organization={DISI, Departement of Computer Science and Engineering, University of Bologna},%Department and Organization
            % addressline={}, 
            city={Bologna},
            % postcode={}, 
            % state={},
            country={Italy}}           
% \affiliation[lamb]{organization={Lamborghini},%Department and Organization
%             % addressline={}, 
%             % city={},
%             % postcode={}, 
%             % state={},
%             country={Italy}}

%% Abstract
\begin{abstract}
%% Text of abstract
This paper introduces Fast Calibrated Explanations, a method designed for generating rapid, uncertainty-aware explanations for machine learning models. By incorporating perturbation techniques from ConformaSight—a global explanation framework—into the core elements of Calibrated Explanations (CE), we achieve significant speedups. These core elements include local feature importance with calibrated predictions, both of which retain uncertainty quantification. While the new method sacrifices a small degree of detail, it excels in computational efficiency, making it ideal for high-stakes, real-time applications. Fast Calibrated Explanations are applicable to probabilistic explanations in classification and thresholded regression tasks, where they provide the likelihood of a target being above or below a user-defined threshold. This approach maintains the versatility of CE for both classification and probabilistic regression, making it suitable for a range of predictive tasks where uncertainty quantification is crucial.
\end{abstract}

%%Graphical abstract
% \begin{graphicalabstract}
% %\includegraphics{grabs}
% \end{graphicalabstract}

%%Research highlights
% \begin{highlights}
%     \item Novelty: Introduces a novel method, Fast Calibrated Explanations, for generating rapid explanations with built-in uncertainty quantification. 
%     \item Significance: Provides a highly efficient approach for estimating feature importance while also quantifying the uncertainty of these estimates.
%     \item Methodology: Leverages perturbation techniques on the calibration set, enhancing the existing Calibrated Explanations framework for speed and efficiency.
%     \item Key Results: Achieves significant speedups in generating calibrated predictions and local feature importance, all with uncertainty quantification.
%     \item Applications: Model-agnostic and applicable to both classification and thresholded regression tasks. Source code is publicly available on GitHub.
% \end{highlights}

%% Keywords
\begin{keyword}
Uncertainty Quantification \sep Calibrated Explanations \sep ConformaSight \sep Explainable AI
%% keywords here, in the form: keyword \sep keyword

%% PACS codes here, in the form: \PACS code \sep code

%% MSC codes here, in the form: \MSC code \sep code
%% or \MSC[2008] code \sep code (2000 is the default)

\end{keyword}

\end{frontmatter}

%% Add \usepackage{lineno} before \begin{document} and uncomment 
%% following line to enable line numbers
%% \linenumbers

%% main text
%%
\newpage
%% Use \section commands to start a section
\section{Introduction}
\label{sec:intro}
%% Labels are used to cross-reference an item using \ref command.

Artificial Intelligence (AI) is becoming an integral part of modern society, influencing everything from retail recommendations to medical diagnosis predictions \citep{das2022artificial,albahri2023systematic} and even defence strategies \citep{devitt2021method}. In many predictive tasks, AI systems are typically developed as models trained via Machine Learning (ML) algorithms. While these models often achieve remarkable accuracy, such as outperforming medical professionals in cancer detection \citep{zou2018ai}, they are neither flawless nor purely objective. Their performance is inherently tied to the data and algorithms used for training, making the outcomes sensitive to these factors.

As AI becomes more ubiquitous, the need for explainability in AI models has grown. In high-stakes applications, such as healthcare or autonomous driving, it is crucial that AI systems provide not only accurate predictions but also transparent reasoning behind those predictions. Explainable AI (XAI) aims to make AI's decision-making processes more understandable to humans, which enhances trust and facilitates informed decision-making by users. Furthermore, explainable models allow developers and stakeholders to identify potential weaknesses, limitations, or unintended consequences of the system. By providing insights into how the AI arrives at its conclusions, XAI helps bridge the gap between complex model operations and human interpretability, ultimately fostering more reliable and accountable AI systems across various domains \citep{gunning19}.

When making decisions based on Machine Learning (ML) models, it is essential to account for the inherent uncertainty in their predictions \citep{romano2020malice,wang2024equal}. While many models provide point estimates, these alone fail to capture the degree of confidence in a given prediction, which can be crucial, particularly in safety-critical applications. Understanding uncertainty helps quantify the reliability of the model's output and aids in making more informed decisions.

Uncertainty in ML models can generally be categorized into two types: aleatoric and epistemic uncertainty \citep{hullermeier2021aleatoric}. Aleatoric uncertainty, also known as statistical or irreducible uncertainty, arises from inherent noise in the data. It represents variability in outcomes due to factors that cannot be explained by the model, such as measurement errors or inherent randomness in the data-generating process. This type of uncertainty cannot be reduced by gathering more data because it is intrinsic to the task.

In contrast, epistemic uncertainty reflects the model's lack of knowledge, typically caused by limited or insufficient data. It is also known as reducible uncertainty because it can be decreased by acquiring more data or improving the model's capacity. Epistemic uncertainty is particularly important in situations where the model is making predictions on out-of-distribution or novel examples—scenarios where the model may be more prone to errors.

In light of these uncertainties, methods like Conformal Prediction (CP) \citep{vovk2005algorithmic} offer a principled framework for uncertainty quantification. CP is a distribution-free, model-agnostic approach that provides reliable confidence intervals for predictions. By not assuming any particular distribution for the data, CP can generate prediction intervals that account for both aleatoric and epistemic uncertainties, making it highly adaptable across different ML tasks and models.

Moreover, for classification tasks or probabilistic estimates, Venn Predictors, an extension of conformal prediction, are particularly useful. Venn Predictors provide a way to output multiple probability estimates. Unlike typical methods that output a single probability estimate (e.g., from softmax layers in neural networks), Venn Predictors yield a set of probabilities that, based on the current available data, ensures one of the probabilities will be correct. This added layer of uncertainty quantification offers more robust probabilistic estimates, which can be crucial in risk-sensitive applications like medical diagnostics or autonomous systems.

The strength of conformal methods, including Venn predictors, lies in their ability to complement standard ML outputs by offering uncertainty estimates that are both mathematically rigorous and practically useful. This enhances the interpretability of AI systems and provides decision-makers with a clearer understanding of the model's confidence, making these methods preferable for applications where understanding uncertainty is as important as the prediction itself.

A natural development has been to integrate conformal methods into explanation techniques, aiming to provide not only accurate predictions but also transparent and trustworthy explanations. Calibrated Explanations \citep{lofstrom2024ce_classification} exemplify this approach by embedding uncertainty directly into the explanation process through calibration. By aligning the confidence of a model's prediction with the reliability of its explanation, this method are offering users a calibrated view of when the model’s reasoning can be trusted. Calibrated Explanations provide local explanations with uncertainty quantification of both prediction and feature importances for individual instances and is applicable to both classification and regression.

Similarly, ConformaSight \citep{yapicioglu2024conformasight} builds on conformal prediction to provide a global explanation framework. It ensures robust explanation sets with guaranteed coverage, making it applicable across different models and robust even in the presence of noisy or perturbed data. This allows for consistent, high-confidence explanations, even in unpredictable environments.

This paper aims to combine the strengths of local explanations from Calibrated Explanations — which maintain applicability across classification and thresholded regression scenarios — with the perturbation approach employed in ConformaSight. Specifically, we preserve the flexibility of explaining predictions for both classification and thresholded regression outputs (e.g., explaining the probability of exceeding a threshold). This is enhanced by incorporating ConformaSight’s perturbation approach, enabling the generation of computationally efficient factual explanations providing feature importance.

% added where fast claibrated explanations could be critically useful - to be added references
The resulting approach offers extremely fast explanations that can be highly advantageous in real-time decision-making processes, particularly where the outputs of these explanations are fed into subsequent steps or systems. Performance is essential for scenarios like machine teaching \citep{goyal2019counterfactual, zhu2018overview}, where explanation algorithms are required to function in real-time, ideally on resource-constrained platforms such as mobile devices. In scenarios requiring critical decisions within a short time frame, such as emergency response or automated monitoring, the ability to generate explanations quickly ensures that the system can continue operating seamlessly. Essentially, these fast explanations become an integral part of a continuous decision-making loop, where real-time insights flow directly into other processes for immediate action. Additionally, this could especially be crucial in time-series or sequential data tasks, where the significance of features may shift dynamically across different time points. In such cases, generating a single static explanation for an entire series would be misleading, as the importance of each feature evolves over time. Hence, fast calibrated explanations can adapt to the changing context at each timestamp to ensure more accurate insights, enhancing the reliability of real-time predictive models.

In addition to delivering rapid explanations with feature importance, it also quantifies uncertainty for both the overall prediction and the individual contribution of each feature. This allows decision-makers to interpret not only the model’s output but also the reliability of each contributing factor, providing deeper insights for risk-sensitive applications where understanding both prediction and explanation uncertainty is critical.

In the next section, thorough descriptions of the building blocks for this paper are provided. In Section~\ref{sec:Proposal}, our contribution is described. The experimental setup is described in Section~\ref{sec:setup} whereas the results provide both evaluation results and a demonstration of its applicability. Section~\ref{sec:conclusion} wraps up the paper with a concluding discussion and pointers for future work. 

\section{Background}
%\citep{liu2022conformalized,calmon2017optimized}

\subsection{Post-Hoc Explanation Methods} \label{sec:Expl_meth}
In machine learning (ML), there are two main approaches to generating explanations. One method involves using inherently interpretable and transparent models, also known as Interpretable AI, which describes the internals of a system in a way that is understandable to humans \citep{Gilpin_explainingExplanations}. Alternatively, post-hoc techniques can be used to explain complex models, making them suitable for explaining black-box models as well \citep{Kenny2021Explaining}.

Post-hoc explanations involve creating simpler models that clarify how a complex model’s predictions are linked to input features. These explanations can be either local, focusing on a single instance, or global, providing insights into the overall model. They often incorporate visual aids like feature importance plots, pixel representations, or word clouds to emphasize the key features, pixels, or words influencing the model’s predictions. The importance of making black-box models explainable is highlighted by \cite{molnar2020interpretable}, which discusses various interpretability methods for machine learning models. Moradi et al. \citep{moradi2021post} propose the Confident Itemsets Explanation (CIE) method, which uses highly correlated feature values to explain black-box classifiers by discretizing the decision space into smaller subspaces. Another interesting approach is introduced by \cite{holzinger2019causability}, which focuses on the quality of explanations in AI systems, particularly in the medical field, by introducing the concept of causability in addition to explainability.

Research on proposed post-hoc methods is diverse and frequently tailored to specific tasks. For instance, \cite{Muddamsetty2022} provides a visual explanation of black-box models by localizing the region responsible for a prediction. This approach is tested on a classification task by localizing entire object classes within an image. %Combined with saliency maps, the perturbation-based method described by \cite{petsiuk2018rise} involves altering the input while monitoring the resulting changes. This approach is tested by feeding the model randomly masked versions of the input image and analysing the corresponding outputs. 
Another popular approach is the use of counterfactual explanations, which, similar to perturbation methods, vary the input space to understand how it affects the output. This approach, employed in a statistical fashion, is used by \citep{Jung2022} to produce human-friendly interpretations on classification tasks.

Post-hoc explanations for classification and regression have some distinguishing characteristics due to the nature of the insights they offer. In classification, explanations involve predicting the class an instance belongs to from predefined classes, with probability estimates reflecting the model’s confidence for each class. Techniques like SHAP \citep{lundberg2017unified}, LIME \citep{Ribeiro2016_kdd}, and Anchor \citep{ribeiro2018anchors} explore factors contributing to class assignment, often using feature importance, such as words in textual data or pixels in images. However, both SHAP and LIME can also be used to generate explanations for regression models. In regression, the focus is on predicting numerical values associated with instances without predefined classes. Explanations for regression models normally adapt techniques designed for classifiers by attributing features to predicted outputs.

Local explanations in classification often rely on probability estimates, which most machine learning models can generate to indicate the likelihood of each class. These probability estimates are commonly interpreted as a measure of prediction confidence. For example, in a binary classification scenario, a model predicting an instance belonging to the positive class with a probability estimate of 0.8 is considered more confident than one predicting the same instance with a probability estimate of 0.65. Probability estimates form the basis for local explanations in classification tasks, where the confidence level indicates the likelihood of the positive class being the true class for that instance.

%%%

\subsection{Calibration and Uncertainty Quantification}
Basing decisions on accurate information is crucial in decision-making, placing an additional layer of requirements on predictive models to provide well-calibrated predictions and guarantees. 

Conformal Prediction (CP) \citep{vovk2005algorithmic} is a distribution-free framework that offers prediction regions with guaranteed coverage, whose value and effectiveness have been demonstrated in numerous studies \citep{Toccaceli2022} \citep{Barber2022ConformalPB}. Errors occur when the true target falls outside the predicted region. However, conformal predictors maintain automatic validity under exchangeability, resulting in an error rate of $\epsilon$ over time. 
Conformal regression (CR) provides prediction intervals with user-decided guaranteed coverage, and conformal predictive systems (CPS) \citep{vovk2017nonparametric} provide a conformal predictive distribution (CPD). The CPDs can be queried for intervals with guaranteed coverage, similar to but more dynamic than CR. This is done by defining intervals based on percentiles in the distribution so that a symmetric interval with 90\% coverage can be achieved using the percentiles $\left[5, 95\right]$. The CPDs can also be queried for the probability of the actual instance value being below a user-given threshold, corresponding to the percentile of the threshold value in the distribution. 

For classification, the focus is generally on the calibration of the probability estimates produced by the classifier, which can be defined as follows: 

\begin{equation}
    \label{calibration}
    p(c \mid p^{c})\approx p^{c},
\end{equation}
where $p^{c}$ represents the probability estimate for a particular class label $c$. This means that a well-calibrated model produces predicted probabilities that match observed accuracy. Consequently, whenever a model assigns a probability estimate of $0.9$ to a label, the accuracy for that label should be approximately $90\%$. 

It is well-known that many predictive models produce poorly calibrated probability estimates \citep{van2019calibration}. An external calibration method can be applied to calibrate a poorly calibrated model using a separate portion of the labelled data, called the calibration set, to adjust the predicted probabilities.

The CP framework defines Venn \citep{vovk2004selfcalibrating} and Venn-Abers (VA) \citep{vovk2014vennabers} predictors that produce multi-probabilistic predictions in the form of confidence-based probability intervals. Venn prediction involves a Venn taxonomy, categorising calibration data for probability estimation. The estimated probability for test instances falling into a category is the relative frequency of each class label among all calibration instances (including the test instance) in that category. Defining a proper Venn taxonomy can be challenging, which is the strength of VA. 

\subsubsection{Venn-Abers Calibration} 
VA Calibration offer automated taxonomy optimisation using isotonic regression, resulting in dynamic probability intervals for binary classification. Since the probability interval includes the well-calibrated probability estimates for the true class label being both negative (lower bound) and positive (upper bound), and the instance must be one or the other, it follows that the interval must contain the true probability. The problem from a predictive perspective is that the true class label is not known. However, the width and location of the interval can provide a lot of information. A smaller interval indicates higher certainty about the prediction, while a larger interval indicates more uncertainty. Since it is often impractical to have only an interval to indicate the probability estimate of the positive class, it is common to use a regularisation of the interval as an estimate for the positive class.  

To define a VA predictor predicting a test object $x_{n+1}$, let $Z = \{z_1,\dots,z_n\}$, where $n=l+q$, be a training set. Each instance $z_i=(x_i,y_i)$ consists of two parts, an object $x_i$ and a target $y_i$. Normally, calibration requires a separate calibration set, motivating a split of the training set into a proper training set $Z_l$ with $l$ instances and a calibration set $Z_q=\{z_1,\dots,z_q\}$\footnote{As we assume random ordering, the calibration set is indexed $1,\dots,q$ rather than $l+1,\dots,n$, for indexing convenience.}. A scoring classifier is trained on $Z_l$ to compute $s$ for $\{x_{1},\dots,x_q,x_{n+1}\}$. The score $s$ is defined as the probability estimate for the positive class from a classifier $h$. Inductive VA prediction follows these steps:

\begin{enumerate}
    \item Use $\{(s_1,y_1),\dots,(s_q,y_q),(s_{n+1},y_{n+1}=0)\}$ to derive the isotonic calibrator $g_0$ and use $\{(s_1,y_1),\dots,(s_q,y_q),(s_{n+1},y_{n+1}=1)\}$ to derive the isotonic calibrator $g_1$.
    \item The probability interval for $y_{n+1}=1$ is defined as $[g_0(s_{n+1}),g_1(s_{n+1})]$ (hereafter referred to as $[p_{low},p_{high}]$, representing the lower and upper bounds of the interval).
    \item The regularised probability estimate for $y_{n+1}=1$, minimising the log loss \citep{vovk2014vennabers}, can be defined as:
    \begin{equation}
        p=\frac{p_{high}}{1-p_{low}+p_{high}}
    \end{equation}
\end{enumerate}

In summary, VA produces a calibrated (regularised) probability estimate $p$ together with a probability interval with a lower and upper bound $[p_{low},p_{high}]$.

\subsubsection{Conformal Predictive Systems}\label{CPS} 
Conformal Predictive Systems produce CPDs for each test object $x_{n+1}$ when the target domain is numeric (i.e. regression). To define a CPS, assume the existence of an underlying regression model $h$ trained using $Z_l$. Like all conformal predictors, CPS relies on nonconformity scores $\alpha$, defining the strangeness of an instance. Unlike CR, where the nonconformity is usually defined as the absolute error $\alpha_i = \left| y_i - h(x_i) \right|$, CPS defines nonconformity using the signed errors $\alpha_i = y_i - h(x_i)$. The prediction for a test instance $x_{n+1}$ then becomes the following CPD:

\begin{equation}
%    \resizebox{.91\linewidth}{!}{$
    \displaystyle
    CPD(y) = 
        \begin{cases}
            \textstyle \frac{i+\tau}{q+1}, \textrm{ if } y\in\left(C_{(i)},C_{(i+1)}\right),	& \textrm{for } i \in \{0,...,q\}\\
            \textstyle \frac{i'-1+(i''-i'+2)\tau}{q+1}, \textrm{if } y = C_{(i)},	& \textrm{for } i\in \{1,...,q\}
        \end{cases}  
%        $}
    \label{eqn:cps-q}
\end{equation}
where $C_{(1)}, \ldots, C_{(q)}$ are obtained from the calibration scores $\alpha_1, \ldots, \alpha_q$, sorted in increasing order:
\begin{equation}
    C_{(i)} = h\left(x_{n+1}\right)+\alpha_i
\end{equation}
with $C_{(0)}=-\infty$ and $C_{(q+1)}=\infty$. In case of a tie, $\tau$ is sampled from the uniform distribution $U(0,1)$, and its role is to allow the $p$-values of target values to be uniformly distributed, $i''$ is the highest index such that $y = C_{(i'')}$, while $i'$ is the lowest index such that $y = C_{(i')}$. 

The following cases provide some further intuition on how a CPD can be used: 
\begin{itemize}
    \item Obtaining a two-sided symmetric prediction interval for a chosen significance level $\epsilon$ can be done by $[C_{\lfloor (\epsilon/2)(q+1) \rfloor}, C_{\lceil (1-\epsilon/2)(q+1) \rceil}]$. Since the CPS has guaranteed coverage, the expected error of the obtained interval will be $\epsilon$ in the long run. Asymmetric prediction intervals are possible by selecting percentiles for the lower ($p^{low}$) and higher ($p^{high}$) bounds of the interval. The guaranteed coverage of the interval will be $\epsilon=p^{high}-p^{low}$.
    \item Still using the significance level $\epsilon$, a lower-bounded one-sided prediction interval can be obtained by $[C_{\lfloor \epsilon(q+1) \rfloor}, \infty]$, and an upper-bounded one-sided prediction interval can be obtained by $[-\infty, C_{\lceil (1-\epsilon)(q+1) \rceil}]$. The coverage guarantees still apply. 
    \item Similarly, a point prediction corresponding to the median of the distribution can be obtained by $(C_{\lceil 0.5(q+1) \rceil}+C_{\lfloor 0.5(q+1) \rfloor})/2$. The median prediction can be seen as a calibration of the underlying model's prediction. Unless the model is biased, the median will tend to be very close to the prediction of the underlying model.
    \item For a specific threshold $t$, the distribution can return the estimated probability $p(C \leq t)$. Thus, it is possible to get the probability of the true target being below the threshold $t$.
\end{itemize}

 A CPS offers richer opportunities to define intervals and probabilities through querying the CPD compared to CR. A particular strength is the ability to calibrate the underlying model. For example, if the underlying model is consistently overly optimistic, the median from the CPS will adjust for that and provide a calibrated prediction that is better adjusted to reality. 

\subsection{Calibrated Explanations}\label{sec:ce}
Calibrated Explanations is a recently released\footnote{Calibrated Explanations can be installed using, e.g., \texttt{pip install calibrated-explanations} or accessed at \href{https://github.com/Moffran/calibrated_explanations}{github.com/Moffran/calibrated\_explanations}.} local explanation method supporting both classification and regression, providing feature importance with uncertainty quantification \citep{lofstrom2024ce_classification,lofstrom2023ce_regression}. Calibrated Explanations produce instance-based explanations, and a \textit{factual explanation} is composed of a \textit{calibrated prediction} from the underlying model accompanied by an \textit{uncertainty interval} and a collection of \textit{factual feature rules}, each composed of a \textit{feature weight with an uncertainty interval} and a \textit{factual condition}, covering that feature's instance value. %It also allows \textit{counterfactual explanations} to provide insights about how changes to one or several features affect the calibrated prediction and uncertainty interval. 
Calibrated Explanations support both binary and multi-class classification. In binary classification, the explanation explains the calibrated probability estimate (and its level of uncertainty) for the positive class, whereas in multi-class classification, the most probable class (after calibration) is considered the positive class and all other classes are treated as the negative class, i.e., not the predicted class. For regression, there are two alternative use cases: 
\begin{enumerate}
    \item The regression explanation explains a calibrated estimate of the prediction from the regressor, with a confidence interval covering the true target with a user-assigned level of confidence.
    \item The thresholded explanation explains the calibrated probability estimate (and its level of uncertainty) for the calibrated estimate of the prediction being below a user-given threshold.
\end{enumerate}

The algorithm's core is agnostic to whether it is a classification or regression problem since it is defined based on a numeric estimate and a lower and an upper bound defining an uncertainty interval for the numeric estimate. For classification, the probability estimate for the positive class is calibrated using a VA calibrator \citep{vovk2014vennabers}, producing a lower and an upper bound for the calibrated probability estimate (using a regularised mean of these bounds as the numeric estimate). For regression, a Conformal Predictive System (CPS) \citep{vovk2020computationally}, producing a Conformal Predictive Distribution (CPD), is used as a calibrator of the underlying model. For the first use case, explaining the prediction value, the numeric estimate is the median from the CPD, and the lower and upper bounds are represented by user-selected percentiles in the CPD, defining the interval with guaranteed coverage. For the second use case, explaining the probability of being below a user-given threshold, the percentile in the CPD representing the threshold position is used as a probability estimate (similar to classification) upon which a VA calibrator is applied. For details on how thresholded regression works, see the original regression paper by \citet{lofstrom2023ce_regression}. 

Calibrated Explanations assume the existence of a predictive model $h$, trained using the proper training set $Z_l$, outputting a numeric value when predicting an object $h(x_i)$. For classification, the model is a scoring classifier, producing probability estimates for the positive class. For regression, it is an ordinary regressor predicting the expected value. Algorithm~\ref{alg:ce} describes how Calibrated Explanations creates a factual explanation of $x$\footnote{The index $n+1$ is omitted to reduce clutter.}. 

% \todo[inline]{Calibrated Explanations}
\begin{algorithm}
\caption{Factual Calibrated Explanations}\label{alg:ce}
\begin{algorithmic}[1]  % The [1] option numbers each line
    \State \textbf{Input:} Fitted model $h$, calibrator, test object $x$
    \State \textbf{Output:} Factual explanation of $x$
    % \State Apply a calibrator on the prediction $h(x)$ to get a calibrated prediction $\varphi$ and uncertainty interval $\left[\varphi_{low},\varphi_{high}\right]$. 
    \If{Classification}
        \State Use VA as calibrator to produce calibrated probability estimate $\varphi=p$ and 
        \Statex \hspace{1.5em}uncertainty interval $\left[\varphi_{low}=p_{low},\varphi_{high}=p_{high}\right]$.
    \EndIf 
    \If{Regression}
        \State Use CPS as calibrator and let $\varphi$ be the median and $\left[\varphi_{low},\varphi_{high}\right]$ is either a 
        \Statex \hspace{1.5em}one- or two-sided interval as described above.
    \EndIf
    \For{each feature $f\in F$}
        \State Changing the value of feature $f$, one at a time in a systematic way, producing 
        \Statex \hspace{1.5em}slightly perturbed versions of object $x$, the calibrator can be used to estimate 
        \Statex \hspace{1.5em}the (averaged) prediction $\varphi_f$ and uncertainty intervals $\left[\varphi_{low\_f},\varphi_{high\_f}\right]$.
        \State The feature importance for feature $f$ is defined as the difference between the 
        \Statex \hspace{1.5em}calibrated prediction $\varphi$, achieved on the original object $x$, and the estimated 
        \Statex \hspace{1.5em}(averaged) calibrated prediction $\varphi_f$, achieved on the perturbed versions of $x$. 
        \State The uncertainty intervals for the feature importance are defined analogously by 
        \Statex \hspace{1.5em}calculating the difference between $\varphi$ and the uncertainty intervals 
        \Statex \hspace{1.5em}$\left[\varphi_{low\_f},\varphi_{high\_f}\right]$ for the perturbed versions of $x$.
        \State A factual feature rule is formed, with a factual condition defined as 
        \Statex \hspace{1.5em}\texttt{feature = categorical instance value}, for categorical features, or 
        \Statex \hspace{1.5em}\texttt{feature $\leq$ threshold} or \texttt{feature $>$ threshold}, for numerical features. 
        \Statex \hspace{1.5em}The \texttt{threshold} is defined so that the factual condition incorporates the 
        \Statex \hspace{1.5em}numerical instance value for that feature. Since the factual condition must 
        \Statex \hspace{1.5em}always include the feature value, only one factual condition is formed for each 
        \Statex \hspace{1.5em}feature.
    \EndFor
    \State \Return A factual explanation, composed of a \textit{calibrated prediction} $\varphi$ from the underlying model accompanied by an \textit{uncertainty interval} $\left[\varphi_{low},\varphi_{high}\right]$ and the collection of \textit{factual feature rules}.
\end{algorithmic}
\end{algorithm}

% \todo[inline]{Use the algorithm environment for all pseudocode and algorithm descriptions}

For further details on the algorithm and how it is applied to classification and the two use cases for regression, see \cite{lofstrom2024ce_classification} and \cite{lofstrom2023ce_regression}.

\subsection{Perturbation Based Explanations}

%\todo[inline]{Describe ConformaSight and the perturbation approach used. Initially, blocks of text is simply copied directly from the paper.}%

\citet{yapicioglu2024conformasight} presents ConformaSight, a unique explanation approach based on conformal prediction methodology that provides insightful and resilient explanations regardless of the underlying data distribution. Its purpose is to give explanations for set-type predictions that conformal predictors create. ConformaSight highlights the influence of the calibration process on prediction outputs, in contrast to conventional explanation approaches that mainly concentrate on feature importance.

In classification tasks, it is essential to understand the factors that influence the formation of prediction sets within conformal prediction frameworks to enhance model interpretability and trust\citep{Vovk:2005:ALRW}. Analyzing metrics such as weighted coverage and weighted set size provides valuable insights into the model’s uncertainty representation \citep{angelopoulos2023conformal}. Weighted coverage measures the proportion of instances that are correctly classified and included within prediction sets, thereby indicating the model’s reliability in identifying uncertain regions relative to class distribution \citep{Shafer2007}. On the other hand, weighted set size offers critical information about the granularity of uncertainty representation, reflecting the average number of instances within prediction sets while accounting for class imbalance. By examining these metrics, researchers can gain a deeper understanding of the relationship between model predictions and input features.
 
Furthermore, ConformaSight focuses on how the calibration process affects prediction intervals, which goes beyond conventional feature importance explanations. A feature in the calibration set is important in influencing the model's confidence in its predictions when it has a significant impact on the coverage of these intervals. This method takes advantage of the finding that, when the conformal prediction algorithm is run multiple times with distinct calibration datasets, the coverage will differ over a limitless number of validation points and yet satisfy the $1 - \alpha$ (error rate) minimum coverage requirement \citep{angelopoulos2021gentle}. This variability highlights the significance of some properties in the calibration set, especially those that have a large impact on the coverage of prediction intervals, suggesting that these factors are critical in determining the model's level of confidence in its predictions.

As a result, ConformaSight provides explanations that highlight the significance of features as well as the calibration process's impact on forecast accuracy. Metrics like \textit{weighted coverage} and \textit{weighted set size} are used to achieve this; they are particularly useful in conformal classification situations \citep{vovk2005algorithmic}, offering a more thorough comprehension of the dynamics of the model.

%In conformal prediction-based models, explanations extend beyond identifying important features; they also reveal how the calibration process affects prediction intervals. For example, if a feature in the calibration set significantly impacts the coverage of prediction intervals, it suggests that the feature plays a key role in shaping the model’s confidence in its predictions. Thus, explaining predictions in ConformaSight involves both highlighting feature importance and detailing how the calibration process impacts predictive performance. This is achieved by leveraging metrics such as \textit{weighted coverage} and \textit{weighted set size} in conformal classification \citep{vovk2005algorithmic}.%

Central to ConformaSight is the idea of leveraging distributional changes within the calibration set to systematically detect variations that highlight feature importance. These changes are introduced through perturbations, mathematically defined in Sections~\ref{def1}, \ref{def2}, and~\ref{def3}.

\subsubsection{Definition 1: Permutation-based Perturbations}\label{def1}
Let the set of categorical features be represented by $C_F$. Let $x_{C_f}$ represent a categorical feature with $c$ different categories for each $C_f \in C_F$. By arbitrarily permuting the values of $x'_{C_f}$ $k$ times, the permutation-based perturbation function \texttt{permute($x_{C_f}$, $k$)} creates a perturbed variant of $x_{C_f}$, called $x'_{C_f}$. Formally, this procedure can be explained as:

\begin{equation} \label{eq:permute} 
    x'_{C_f} = \texttt{permute}(x_{C_f}, k) 
\end{equation}

%\subsubsection{Definition 1: Permutation-based Perturbations}
%Let $C_F$ be the set of categorical features. For each $C_f\in C_F$, let %$x_{C_f}$ represent a categorical feature with $c$ unique categories. The %permutation-based perturbation function, \texttt{permute($x_{C_f}$,$k$)}, %generates a perturbed version of $x_{C_f}$, denoted as $x´_{C_f}$, where the %values of $x'_{C_f}$ are randomly permuted $k$ times. Formally, this can be %written as:

%\subsubsection{Definition 2: Gaussian Noise Perturbations}
%Let $N_F$ be the set of numerical features. For each $N_f\in N_F$, let $x_{N_f}$ %be a numerical feature with standard deviation $\sigma$. Given a severity %parameter $s$, the Gaussian noise perturbation function adds noise sampled from %a normal distribution to $x_{N_f}$, resulting in a perturbed feature $x'_{N_f}$. %Mathematically, this is expressed as:%

\subsubsection{Definition 2: Gaussian Noise Perturbations}\label{def2}
Define the set of numerical characteristics by $N_F$. Consider a numerical feature $x_{N_f}$ with a standard deviation $\sigma$ for each $N_f \in N_F$. The Gaussian noise perturbation algorithm takes a severity parameter $s$ and adds noise to $x_{N_f}$, sampled from a normal distribution; this produces the perturbed feature $x'_{N_f}$. Theoretically this procedure is defined as:

\begin{equation} \label{eq:normal} 
    x'_{N_f} = x_{N_f} + \eta(s), \quad \text{where } \eta(s) \sim \texttt{Normal}(0, s \times \sigma) 
\end{equation}

\subsubsection{Definition 3: Uniform Noise Perturbations}\label{def3}
The uniform noise perturbation function adds noise from a uniform distribution to a numerical feature \(x_{N_f}\). The definition of the perturbed feature \(x'_{N_f}\) given a severity parameter \(s\) is as outlined below:

\begin{equation} \label{eq:gaussian}
    x'_{N_f} = x_{N_f} + \eta(s), \quad \text{where } \eta(s) \sim \texttt{Uniform}(-s \times R_{N_f}, s \times R_{N_f}) 
\end{equation}

where $R_{N_f}$ represents the range of values in $x_{N_f}$. The uniform noise perturbation introduces variability across the dataset, facilitating the exploration of various distributional shifts.

By implementing these perturbations on a feature-by-feature basis, ConformaSight assesses the changes in the model's behavior relative to the original calibration set. This alteration acts as a measure of each feature's significance, offering enhanced insights into how individual features impact the model's predictions.

%By applying these perturbations on a feature-by-feature basis, ConformaSight evaluates the relative change in the model's behavior compared to the original calibration set. This change serves as an indicator of the relative importance of each feature, providing deeper insights into how individual features influence the model's predictions.%

% \todo[inline]{The description above needs re-writing and may also be extended depending on where we decide to submit. I do not think we need all the algorithms etc, but maybe some additional details may be necessary or convenient.

% One of the things we need to make sure is that we use a consistent notation and terminology. My suggestion is that we use the Calibrated Explanations notation, as it covers more text, thus saving us some work. :)}

\section{Proposed Solution}\label{sec:Proposal}
% Provide an introduction that fits the background and related work. 
This paper aims to incorporate the perturbation approach, constituting a core element in ConformaSight, into the explanation method Calibrated Explanations, making it possible to extract a new form of explanations, providing fast feature importance generation without rule conditions. 

% \citep{liu2022conformalized,calmon2017optimized}.
As described in~\ref{sec:ce}, perturbations are done for each feature of the test instance at explanation time in Calibrated Explanations. The solution that we propose in this paper performs all perturbations on the calibration set at initialisation of the Fast Calibrated Explanations, resulting in some additional overhead once when initialising Fast Calibrated Explanations, while avoiding any perturbations at explanation time. Compared to the explanations in Calibrated Explanations, a main difference with the solution proposed here is that there is no rule conditions. Instead, each feature is assigned a feature weight determining the relative importance of that feature compared to other features. As such, the provided explanations are factual, conveying the feature importance per instance. The resulting solution provides very fast explanations with feature weights that can be analysed per instance.

More formally, the solution that we propose can be divided into two stages: 1) initialisation, and 2) explanation. The initialisation stage 1) is described in Algorithm~\ref{alg:init} while the explanation stage 2) is described in Algorithm~\ref{alg:explanation}.

% \todo[inline]{Describe Alg init and alg explanation.}
\begin{algorithm}
\caption{Initialisation of Fast Calibrated Explainer}\label{alg:init}
\begin{algorithmic}[1]  % The [1] option numbers each line
    \State \textbf{Input:} Calibration set $Z_q$, factor $k$, severity $s$, noise $\eta$
    \State \textbf{Output:} An initialised Fast Calibrated Explainer
    \State Multiply $Z_q$ with factor $k$ resulting in $\mathbf{Z}_q=[Z_q]_{i=1}^k$
    \For{each feature $f\in F$}
        \State Permute the $k$ copies of $X_f$ to create a permuted $\mathbf{X}'_f$ using Equation~\eqref{eq:permute}, \eqref{eq:normal}, 
        \Statex \hspace{1.5em}or \eqref{eq:gaussian}.
        \State Initiate a calibrator $\mathcal{C}_f$ using the multiplied calibration set $\mathbf{Z}_q$, substituting the 
        \Statex \hspace{1.5em}$k$ original $X_f$ with $\mathbf{X}'_f$. The kind of calibrator is either a VA for classification or 
        \Statex \hspace{1.5em}a combination of CPS and VA for thresholded regression explanations.
    \EndFor
    \State Initiate a base calibrator $\mathcal{C}$ using the original calibration set $Z_q$
    \State \Return A new Fast Calibrated Explainer with all permutations and calibrators stored
\end{algorithmic}
\end{algorithm}

\begin{algorithm}
\caption{Explanation using Fast Calibrated Explainer}\label{alg:explanation}
\begin{algorithmic}[1]  % The [1] option numbers each line
    \State \textbf{Input:} An initialised Fast Calibrated Explainer, a test object $x$
    \State \textbf{Output:} An explanation of $x$    
    \State Use the base calibrator $\mathcal{C}$ to produce calibrated estimate $\varphi$ and uncertainty interval $\left[\varphi_{low},\varphi_{high}\right]$ for $x$.
    \For{each feature $f\in F$}
        \State Estimate the prediction $\varphi_f$ and uncertainty intervals $\left[\varphi_{low\_f},\varphi_{high\_f}\right]$ using $\mathcal{C}_f$.
        \State The feature weights for feature $f$ is defined as the difference between the 
        \Statex \hspace{1.5em}calibrated prediction $\varphi$, achieved on the original object $x$, and the estimated 
        \Statex \hspace{1.5em}(averaged) calibrated prediction $\varphi_f$. 
        \State The uncertainty intervals for the feature weights are defined analogously by 
        \Statex \hspace{1.5em}calculating the difference between $\varphi$ and the uncertainty intervals 
        \Statex \hspace{1.5em}$\left[\varphi_{low\_f},\varphi_{high\_f}\right]$.
    \EndFor
    \State \Return An explanation, composed of a \textit{calibrated prediction} $\varphi$ from the underlying model accompanied by an \textit{uncertainty interval} $\left[\varphi_{low},\varphi_{high}\right]$ and the collection of \textit{feature weights} with \textit{feature weight uncertainties}.
\end{algorithmic}
\end{algorithm}

Since the perturbation is performed at initialisation and is using a CPD when used for regression, all explanations from a single model that use the same uncertainty interval, i.e., the same percentiles, will result feature weights that have exactly the same size in relation to all other feature weights. The only thing that will differ from instance to instance is the scale of the feature weights, making Fast Calibrated Explanations for standard regression clearly less useful. The same issue does not exist for probabilistic explanations (applicable to both classification and thresholded regression).

The similarities and differences between Calibrated Explanations and Fast Calibrated Explanations are summarised in Table~\ref{tab:comparison}.
\begin{table}[htpb]
    \centering
    \footnotesize
    \caption{Comparison between Calibrated Explanations and Fast Calibrated Explanations}
    \begin{tabular}{l|p{4.5cm}|p{4.5cm}}
        \multicolumn{3}{c}{}\\
       & Calibrated Explanations & Fast Calibrated Explanations \\
       \hline
      Perturbation & Perturbation is done at explanation time on the test instance & Perturbation is done at initialisation time on the calibration set, requiring additional memory space for the perturbed calibration set \\
      \hdashline
      Expressiveness & Each rule contains a condition for which the feature weight applies & No condition is used, with the implication that the feature weight is less expressive \\
      \hdashline
      Interpretation & Each rule condition clearly convey when the feature weight apply, providing clear cues for interpretation & The feature weights provide insights on how much and in which direction a feature affects the prediction, with positive weights favouring the positive class and vice versa \\
      \hdashline
      Regression & Supports both standard regression and thresholded regression & Is only clearly useful for thresholded regression \\
      \hdashline
      Uncertainty & Uncertainty quantification is provided for both prediction and feature weights & Uncertainty quantification is provided for both prediction and feature weights \\
      \hdashline
      Alternatives & Alternative explanations, indicating what prediction the model would output if the feature is altered in accordance with the condition, can be extracted & No alternative explanations can be extracted \\
      \hdashline
      Conjunctions & Conjunctive explanations, indicating the joint impact from several conditional rules, are possible & No conjunctions are available \\
      \hdashline
      Speed & Since perturbations are done on the test instance and since a suitable rule condition must be identified, the explanation time is penalised & Since all perturbations have already been done on the calibration set and no condition is used, explanations can be generated much faster \\
      \hline
    \end{tabular}
    \label{tab:comparison}
\end{table}

\section{Experimental Setup}\label{sec:setup}
The evaluation is divided into two parts. The first part contains a comparative evaluation between the Fast Calibrated Explanations, our proposed solution, and Calibrated Explanations on classification and regression problems. The evaluation is performed separately for classification and regression, focusing on complementary aspects. For code, see the \href{https://github.com/Moffran/calibrated_explanations/blob/main/evaluation/FastCE}{calibrated\_explanations/evaluation/FastCE} folder in the repository. 

For classification, an ablation study of the impact from possible permutation parameters is performed. The evaluation covers both computational cost and how the mean variance of feature weights vary across different parameter settings. The ablation study includes various parameter values for the Fast Calibrated Explanations including both forms of noise type (\textit{uniform} and \textit{gaussian}), four different scaling factors ($1,3,5,10$), and five different severity values ($0,0.25,0.5,0.75,1$). Computational time and mean variance per feature importance is reported for $25$ binary classification data sets. Each data set was split into $50\%$ training data, $25\%$ calibration data and $25\%$ test data and the underlying model was a \texttt{RandomForestClassifier} with 100 trees\footnote{To run the classification experiment, run first \textit{Perturbation\_Experiment\_Ablation.py} to generate results and then run the notebook \textit{Perturbation\_Analysis\_Times.ipynb}.}.

For regression, focus is on computational speed, stability and robustness in comparison with SHAP and LIME applied on calibrated models (using the CPS as calibrator of the underlying model). All target values were min-max normalised to the range $[0,1]$. Each data set was split using $200$ calibration instances, $100$ test instances, and the remaining instances as training set. For each one of the $31$ data sets, six different setups where evaluated, three for standard regression and two for thresholded regression (using $0.5$ as threshold). Two setups applied SHAP and LIME on a calibrated model, two setups used Calibrated Explanations for standard and thresholded regression and two setups used Fast Calibrated Explanations for standard and thresholded regression. Fast Calibrated Explanations is evaluated for standard regression, to indicate the potential for that kind of context, even though these explanations are deemed less useful. The underlying model was a \texttt{RandomForestRegressor} which was calibrated using a CPS, either explicitly (for LIME and SHAP) using the \texttt{ConformalPredictiveSystem} class from \textit{crepes} or implicitly as part of using the \texttt{CalibratedExplainer} class\footnote{To run the regression experiment, run first \textit{Regression\_Experiment.py} to generate results and then run the notebook \textit{Regression\_Analysis.ipynb}.}. The following metrics are evaluated:
\begin{itemize}    
    \item \textit{Stability} means that multiple runs on the same instance and model should produce consistent results. Stability is evaluated by generating explanations for the same predicted instances 5 times with different random seeds (using the iteration counter as random seed). The random seed is used to initialise the \texttt{numpy.random.seed()} and by the discretizers. 
    The largest variance in feature weight (or feature prediction estimate) can be expected among the most important features (by definition of having higher absolute weights). 
    The top feature for each test instance is identified as the feature being most important most often in the 100 runs (i.e., the mode of the feature ranks defined by the absolute feature weight). The variance for the top feature is measured over the 100 runs and the mean variance among the test instances is reported. 
    \item \textit{Robustness} means that small variations in the input should not result in large variations in the explanations. Robustness is measured in a similar way as stability, but with the training and calibration set being randomly drawn and a new model being fitted for each run, creating a natural variation in the predictions of the same instances without having to construct artificial instances. Again, the variance of the top feature is used to measure robustness. The same setups as for stability are used except that each run use a new model and calibration set and that the random seed was set to $42$ in all experiments.    
    \item \textit{Explanation time} is compared between the setups regarding explanation generation times (in seconds per instance). It is only the method call resulting in an explanation that is measured. Any overhead in initiating the explainer class is not considered).  
\end{itemize}

\begin{table}[htbp!]
    \centering
    \footnotesize
    \caption{Explanation time in seconds per instance for binary classification datasets}
    % The results from re-running the experiment, making sure the initialisation was done prior to explanation (as opposed to the previous results)
    \begin{tabular}{lc|cc}
        \multicolumn{4}{c}{}\\
        \textbf{Dataset} & \multicolumn{1}{c|}{\textbf{\#Features}} & \multicolumn{1}{c}{\textbf{CE}} & \multicolumn{1}{c}{\textbf{FCE}}\\
        \hline
        colic & 60 & .369 & .004 \\
        creditA & 43 & .594 & .002 \\
        diabetes & 9 & .034 & .000 \\
        german & 28 & .210 & .001 \\
        haberman & 4 & .004 & .001 \\
        heartC & 23 & .058 & .002 \\
        heartH & 21 & .045 & .002 \\
        heartS & 14 & .032 & .001 \\
        hepati & 20 & .034 & .003 \\
        iono & 34 & .236 & .003 \\
        je4042 & 9 & .021 & .001 \\
        je4243 & 9 & .028 & .001 \\
        kc1 & 22 & .340 & .001 \\
        kc2 & 22 & .121 & .002 \\
        kc3 & 40 & .475 & .003 \\
        liver & 7 & .011 & .001 \\
        pc1req & 9 & .009 & .002 \\
        pc4 & 38 & 1.336 & .002 \\
        sonar & 61 & .833 & .007 \\
        spect & 23 & .032 & .003 \\
        spectf & 45 & .421 & .004 \\
        transfusion & 5 & .008 & .000 \\
        ttt & 28 & .168 & .001 \\
        vote & 17 & .045 & .001 \\
        wbc & 10 & .038 & .001 \\
        \hline
        \textbf{Mean} & \textbf{24.0} & \textbf{.220} & \textbf{.002} \\
    \end{tabular}
    \label{tab:expl_time}
\end{table}

\section{Experimental Results} \label{sec:results}
\subsection{Performance Evaluation for Classification and Regression}
\subsubsection{Comparison of Computation Time}
The code used for the evaluation of Calibrated Explanations (CE) and Fast Calibrated Explanations (FCE) can be found in the \href{https://github.com/Moffran/calibrated_explanations/blob/main/evaluation/FastCE}{FastCE} folder in the repository. As the difference in initialisation time is almost negligible, the results are not presented here but can be found in the folder above. The explanation time clearly differs between Calibrated Explanations (CE) and Fast Calibrated Explanations (FCE). However, as the explanation time between different parameter settings for FCE does not differ much, only the default parameters (noise type=\textit{uniform}, scale factor=5, and severity=0.5) are compared with CE in Table~\ref{tab:expl_time}. 

\begin{table}[htbp]
    \centering
    \footnotesize
    \caption{Average explanation speedup factor for Fast Calibrated Explanations compared to Calibrated Explanations for different parameter settings}
    % The results from re-running the experiment, making sure the initialisation was done prior to explanation (as opposed to the previous results)
    \begin{tabular}{l|cccc|cccc|c}
        \multicolumn{10}{c}{}\\
        \textbf{Noise Type} & \multicolumn{4}{c|}{\textbf{gaussian}} & \multicolumn{4}{c|}{\textbf{uniform}} &  \\
        \textbf{Scale Factor} & \textbf{1} & \textbf{3} & \textbf{5} & \textbf{10} & \textbf{1} & \textbf{3} & \textbf{5} & \textbf{10} &  \\
        \textbf{Severity} &       &       &       &       &       &       &       &       & \multicolumn{1}{c}{\textbf{Mean}} \\
        \hline
        \multicolumn{1}{c|}{\textbf{0}} & 114 & 105 & 110 & 113 & 120 & 110 & 107 & 115 & \textbf{112} \\
        \multicolumn{1}{c|}{\textbf{0.25}} & 96 & 110 & 112 & 118 & 103 & 109 & 110 & 114 & \textbf{109} \\
        \multicolumn{1}{c|}{\textbf{0.5}} & 94 & 115 & 112 & 114 & 110 & 116 & \underline{112} & 117 & \textbf{111} \\
        \multicolumn{1}{c|}{\textbf{0.75}} & 114 & 108 & 116 & 112 & 109 & 114 & 109 & 117 & \textbf{113} \\
        \multicolumn{1}{c|}{\textbf{1}} & 116 & 109 & 114 & 113 & 115 & 116 & 115 & 110 & \textbf{114} \\
        \hline
        \textbf{Mean} & \textbf{107} & \textbf{109} & \textbf{113} & \textbf{114} & \textbf{111} & \textbf{113} & \textbf{111} & \textbf{115} & \textbf{112} \\
    \end{tabular}%
    \label{tab:speedup}
\end{table}

As can be seen, the explanation time in seconds per instance is on average more than $100$ times higher for CE compared to FCE. It is worth noting that the average explanation time for CE is heavily influenced by a few really costly data sets. The median explanation time for CE is $0.045$, being only about $30$ times higher for CE compared to FCE. The average speedup across the various evaluated setups is shown in Table~\ref{tab:speedup} (with the underlined result representing the result in Table~\ref{tab:expl_time}). The average speedup is similar across the various evaluated settings, with some minor deviations. 

\begin{table}[htbp!]
    \centering
    \footnotesize
    \caption{Average initialisation computational cost factor for Fast Calibrated Explanations compared to Calibrated Explanations for different parameter settings}
    \begin{tabular}{l|cccc|cccc|c}
        \multicolumn{10}{c}{}\\
        \textbf{Noise Type} & \multicolumn{4}{c|}{\textbf{gaussian}} & \multicolumn{4}{c|}{\textbf{uniform}} &  \\
        \textbf{Scale Factor} & \textbf{1} & \textbf{3} & \textbf{5} & \textbf{10} & \textbf{1} & \textbf{3} & \textbf{5} & \textbf{10} &  \\
        \textbf{Severity} &       &       &       &       &       &       &       &       & \multicolumn{1}{c}{\textbf{Mean}} \\
        \hline
        \multicolumn{1}{c|}{\textbf{0}} & 20.9 & 21.9 & 23.2 & 26.7 & 20.3 & 22.4 & 23.2 & 26.5 & \textbf{23.1} \\
        \multicolumn{1}{c|}{\textbf{0.25}} & 24.1 & 21.7 & 24.1 & 26.7 & 21.1 & 23.1 & 23.9 & 27.4 & \textbf{24.0} \\
        \multicolumn{1}{c|}{\textbf{0.5}} & 21.0 & 21.9 & 23.8 & 27.7 & 21.7 & 22.4 & \underline{24.8} & 27.0 & \textbf{23.8} \\
        \multicolumn{1}{c|}{\textbf{0.75}} & 20.2 & 22.2 & 24.4 & 27.4 & 20.7 & 22.7 & 24.8 & 26.9 & \textbf{23.7} \\
        \multicolumn{1}{c|}{\textbf{1}} & 19.7 & 22.1 & 23.3 & 27.3 & 19.8 & 22.5 & 24.2 & 27.7 & \textbf{23.3} \\
        \hline
        \textbf{Mean} & \textbf{21.2} & \textbf{22.0} & \textbf{23.8} & \textbf{27.2} & \textbf{20.7} & \textbf{22.6} & \textbf{24.2} & \textbf{27.1} & \textbf{23.6} \\
    \end{tabular}
    \label{tab:init}
\end{table}

So, what is the trade-off that we need to make to get this speedup? Table~\ref{tab:init} show the computational cost in terms of how many times faster the initialisation of Calibrated Explanations is compared to Fast Calibrated Explanations. The average initialisation time for Calibrated Explanations is $8.3e-5$, as a comparison.

On average, the one-time cost for initialisation is that it takes about $25$ times longer for Fast Calibrated Explanations compared to Calibrated Explanations. It means that for the default setup, the average initialisation time for Fast Calibrated Explanations across all data sets was $0.0013$ seconds, whereas the average explanation time per instance across all data sets is $0.0020$ seconds. So even if the initialisation time for Fast Calibrated Explanations is substantially larger than for Calibrated Explanations, it is still negligible considering it is only done once. 

In the evaluation for regression, a comparison is made with calibrated SHAP and LIME. The explanation time in seconds per instance is tabulated in Table~\ref{tab:reg_time}.

\begin{table}[htbp]
    \centering
    \footnotesize
    \caption{Explanation time in seconds per instance for regression datasets}
    \begin{tabular}{lcccc|cc}
        \multicolumn{7}{c}{}\\
        \textbf{Dataset} & \textbf{LIME} & \textbf{SHAP} & \textbf{CE} & \textbf{FCE} & \textbf{PCE} & \textbf{PFCE} \\
        \hline
        abalone & .071 & .239 & .021 & .001 & .022 & .003 \\
        anacalt & .044 & .044 & .014 & .001 & .015 & .002 \\
        bank8fh & .109 & .267 & .022 & .002 & .024 & .004 \\
        bank8fm & .102 & .252 & .021 & .001 & .023 & .004 \\
        bank8nh & .108 & .269 & .022 & .001 & .023 & .003 \\
        bank8nm & .107 & .265 & .022 & .001 & .023 & .003 \\
        comp & .093 & .632 & .041 & .002 & .042 & .004 \\
        concreate & .069 & .180 & .020 & .001 & .021 & .003 \\
        cooling & .071 & .155 & .019 & .001 & .019 & .002 \\
        deltaA & .114 & .052 & .011 & .001 & .013 & .002 \\
        deltaE & .141 & .070 & .014 & .001 & .015 & .003 \\
        friedm & .098 & .042 & .010 & .001 & .012 & .002 \\
        heating & .071 & .154 & .019 & .001 & .020 & .002 \\
        kin8fh & .113 & .313 & .024 & .002 & .027 & .004 \\
        kin8fm & .117 & .326 & .025 & .002 & .029 & .004 \\
        kin8nh & .116 & .320 & .024 & .002 & .026 & .004 \\
        kin8nm & .116 & .334 & .025 & .002 & .025 & .005 \\
        laser & .083 & .027 & .008 & .001 & .008 & .002 \\
        mg & .108 & .071 & .014 & .001 & .015 & .002 \\
        mortage & .076 & .459 & .064 & .002 & .066 & .003 \\
        plastic & .063 & .016 & .003 & .000 & .004 & .001 \\
        puma8fh & .115 & .291 & .023 & .001 & .024 & .004 \\
        puma8fm & .108 & .265 & .022 & .002 & .025 & .004 \\
        puma8nh & .108 & .276 & .022 & .002 & .025 & .004 \\
        puma8nm & .108 & .271 & .022 & .002 & .024 & .004 \\
        quakes & .077 & .023 & .005 & .001 & .006 & .001 \\
        stock & .067 & .361 & .026 & .001 & .027 & .003 \\
        treasury & .075 & .448 & .063 & .001 & .065 & .003 \\
        wineRed & .075 & .485 & .036 & .001 & .037 & .003 \\
        wineWhite & .087 & .579 & .035 & .001 & .037 & .004 \\
        wizmir & .073 & .231 & .026 & .001 & .028 & .003 \\
        \hline
        \textbf{Mean} & \textbf{.093} & \textbf{.249} & \textbf{.023} & \textbf{.001} & \textbf{.025} & \textbf{.003} \\
    \end{tabular}
  \label{tab:reg_time}%
\end{table}%
The speedup is not as impressive for regression as it was for classification, but it is still substantial. FCE is almost $19$ times faster for standard regression (CE) and Probabilistic Fast Calibrated Explanations (PFCE) is more than $8$ times faster than Probabilistic Calibrated Explanations (PCE). FCE is more than $75$ times faster than calibrated LIME and more than $200$ times faster than calibrated SHAP. The relatively smaller speedup observed in regression can be attributed to the shorter baseline explanation times for CE in regression tasks when compared to classification scenarios. 

\subsubsection{Stability and Robustness}
Both stability and robustness have been evaluated for the regression data. The mean and median stability and robustness aggregated over all regression data sets are shown in Table~\ref{tab:stab_rob}. The motivation for using both aggregation methods is that some data sets deviated drastically from the general picture for some methods, having a huge impact on the mean but not on the median. For extremely low variation (less than $1e-30$), the value was set to 0. Detailed result per data set can be found in the evaluation folder. The robustness must be compared to the amount of variability in the predictions from the underlying model, which was $1.7e-4$ on average.

\begin{table}[htbp]
    \centering
    \footnotesize
    \caption{Mean and median stability and robustness aggregated over all regression data sets.}
    % Table generated by Excel2LaTeX from sheet 'Table_stab_rob'
    \begin{tabular}{llcccc|cc}
          & & \textbf{LIME} & \textbf{SHAP} & \textbf{CE} & \textbf{FCE} & \textbf{PCE} & \textbf{PFCE} \\
        \hline
        \textbf{Stability} & Mean & 1.3e-5 & 1.6e-7 & 0 & 1.6e-5 & 4.8e-3 & 4.2e-3 \\
        & Median & 8.8e-6 & 0 & 0 & 6.0e-6 & 3.9e-3 & 3.8e-3 \\
        \hline
        \textbf{Robustness} & Mean & 1.1e-3 & 2.5e-4 & 7.1e-3 & 4.8e-5 & 2.3e-2 & 6.7e-3 \\
        & Median & 5.6e-4 & 1.1e-4 & 2.9e-3 & 3.2e-5 & 1.9e-2 & 6.4e-3 \\
    \end{tabular}%
    \label{tab:stab_rob}
\end{table}

Lets start by considering the standard regression results from LIME, SHAP, CE and FCE, which can all be compared. Considering the results reported in \cite{lofstrom2023ce_regression}, the results for LIME, SHAP and CE are as expected. It is clear that stability is worse for FCE than for CE and SHAP but comparable to LIME. Furthermore, the robustness of FCE is even better than all of the other three methods. Looking at the probabilistic results, it does not make sense to compare these results with the previously mentioned results, as their predictions are probabilities. Both PCE and PFCE are less stable and robust and the reason is related to the sensitivity of the probabilities derived from the CPD. The reason for the sensitivity is that a relatively small change in prediction can easily result in a comparably much larger change in probability for exceeding the threshold, especially if the target is close to the threshold (which is set to $0.5$, i.e., the mid-point in the interval of possible target values). Results are comparable between PCE and PFCE.

\begin{figure}[htbp!]
    \begin{lstlisting}
from calibrated_explanations import WrapCalibratedExplainer
# Load and pre-process your data
# Divide it into proper training, calibration, and test sets

# Initialize the WrapCalibratedExplainer with your model
classifier = WrapCalibratedExplainer(ClassifierOfYourChoice())
regressor = WrapCalibratedExplainer(RegressorOfYourChoice())

# Train your model using the proper training set
classifier.fit(X_proper_training_cl, y_proper_training_cl)
regressor.fit(X_proper_training_reg, y_proper_training_reg)

# Calibrate your model using the calibration set 
# Ensure fast initialisation by assigning fast=True
classifier.calibrate(X_calibration_cl, y_calibration_cl, fast=True)
regressor.calibrate(X_calibration_reg, y_calibration_reg, fast=True)

# Create and plot fast explanations for classification
explanations = classifier.explain_fast(X_test_cl)
explanations.plot(uncertainty=True)

# Create and plot fast explanations for thresholded regression
my_threshold = 500
explanations = regressor.explain_fast(X_test_reg, 
                                    threshold=my_threshold) 
explanations.plot(uncertainty=True)
    \end{lstlisting}
    \caption{Code example on using \textit{calibrated-explanations} for fast explanations}
    \label{lst:code}
\end{figure}

\subsection{Demonstration}
In the demonstration below, a few different data sets are included, to show examples from different use cases. The examples will illustrate how the explanations may look and how they can be understood. The demonstration will include both binary and multi-class data sets, as well as a regression data set for which a thresholded regression explanation is given. Since the calibrated probability of the test instances can be retrieved by the explainer and very certain predictions without much uncertainty will generally not prompt further investigation, all the examples are for less certain predictions. In order to produce an explanation plot conveying uncertainty, code similar to the example in Figure~\ref{lst:code} can be used. 

\begin{figure}[htbp!]
    \centering
    \includegraphics[width=0.8\linewidth]{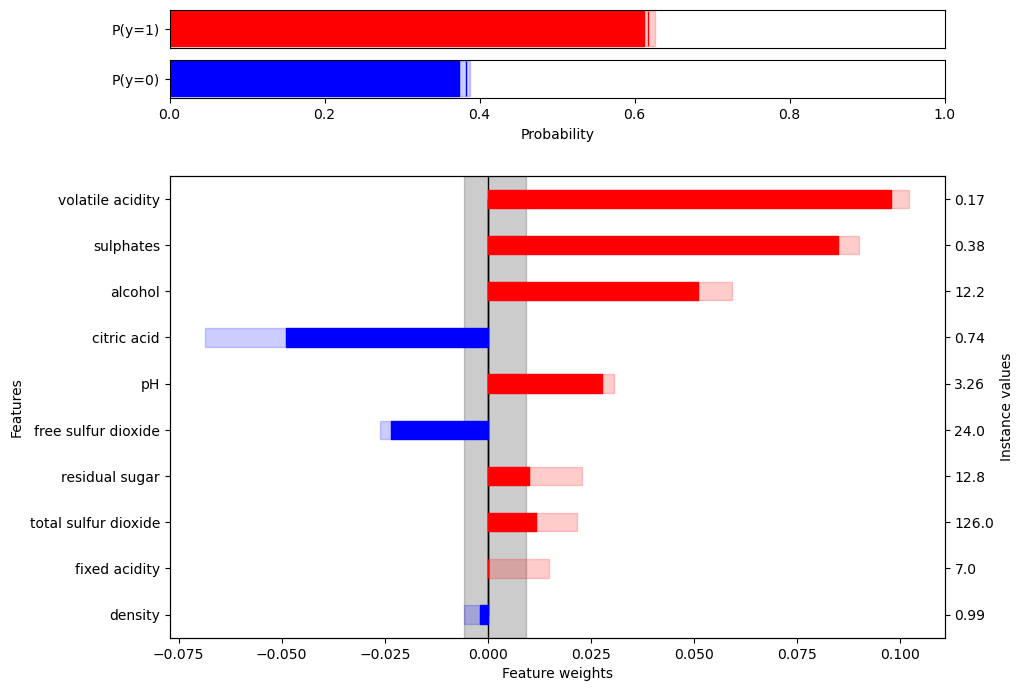}
    \caption{A fast explanation for an instance from the Wine data set}
    \label{fig:wine171}
\end{figure}

To begin the demonstration, the binary classification \textit{wine} data set is used. Figure~\ref{fig:wine171} shows an instance which is predicted as likely to be positive (indicated by the red bar at the top). The plot box at the lower part of the figure provides the contribution by each of the features to the prediction. To the right of the plot box, the feature names are written and to the right of the plot box, the actual feature values of the instance are shown. The interpretation is that the instance value for \textit{Volatile Acidity} make the probability for the positive class be higher, and the instance value for \textit{Citric Acid} make the probability for the positive class be slightly lower, and so on. Each of the bars represent the importance of a particular feature and is indicative of that features importance for this instance. The grey area in the background correspond to the light red area at the top, indicating the uncertainty interval of the calibrated probability for the positive class. The lighter red or blue areas on each feature weight bar indicate the uncertainty interval for the feature weight. This particular instance is in fact class $1$, which several of the features indicate. 

\begin{figure}
    \centering
    \includegraphics[width=0.8\linewidth]{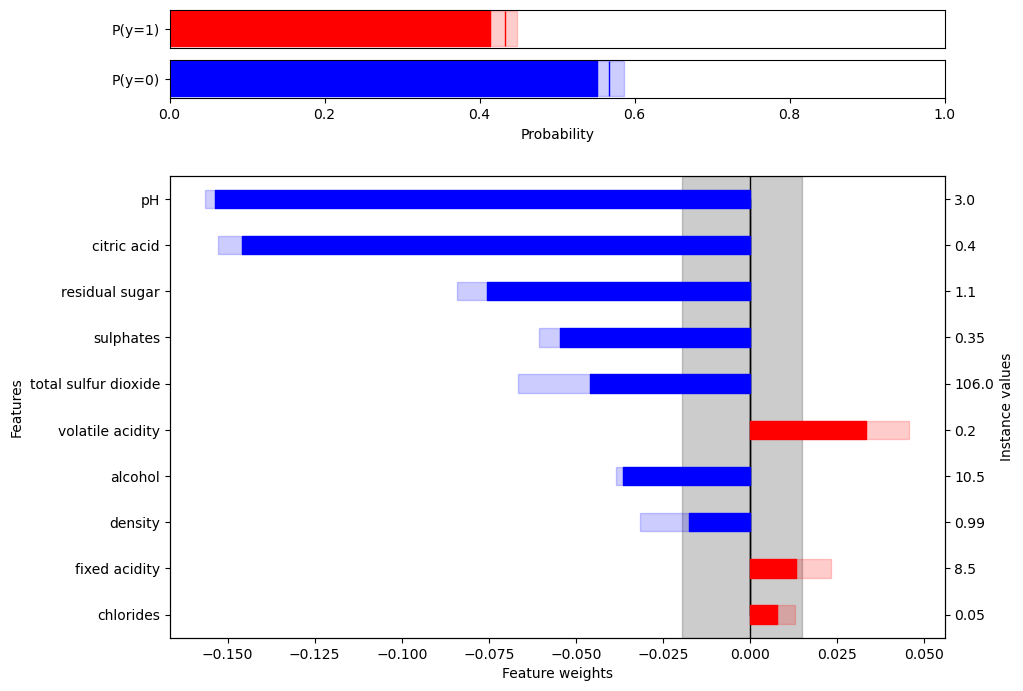}
    \caption{A fast explanation for another instance from the Wine data set}
    \label{fig:wine74}
\end{figure}

Another example for the same data set can be seen in Figure~\ref{fig:wine74}. In this example, the presence of several features each contribute (individually) to lower probability for the positive class (indicated by negative weights, which is shown as blue bars). This particular instance is in fact class $0$.

\begin{figure}
    \centering
    \includegraphics[width=0.8\linewidth]{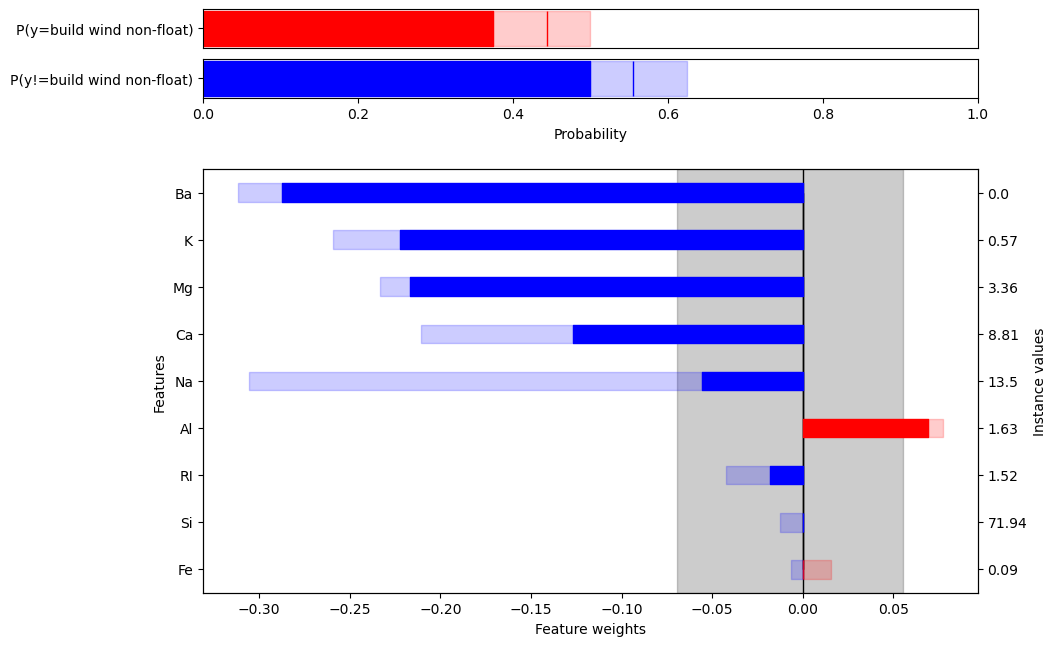}
    \caption{A fast explanation for an instance from the Glass data set}
    \label{fig:glass}
\end{figure}

Looking instead at an example from the multi-class \textit{Glass} data set, the explanation provides the probability for the class being most likely after calibration. As is exemplified in the paper introducing multi-class Calibrated Explanations \citep{lofstrom2024ce_multiclass}, it is advisable to compare the explanations against a confusion matrix, to get an understanding of the typical errors made by the model. Here, it suffices to say that both precision and recall for the class \textit{build wind non-float} is just above $0.7$. In this case, as seen in Figure~\ref{fig:glass}, several of the features are blue, indicating that they contribute to lower probability for the predicted class, strengthening the indication provided by the probability against the predicted class. One of the feature weights, for \textit{Na}, has a large degree of uncertainty, indicating that the weight can be either rather low or rather high. 

\begin{figure}
    \centering
    \includegraphics[width=0.8\linewidth]{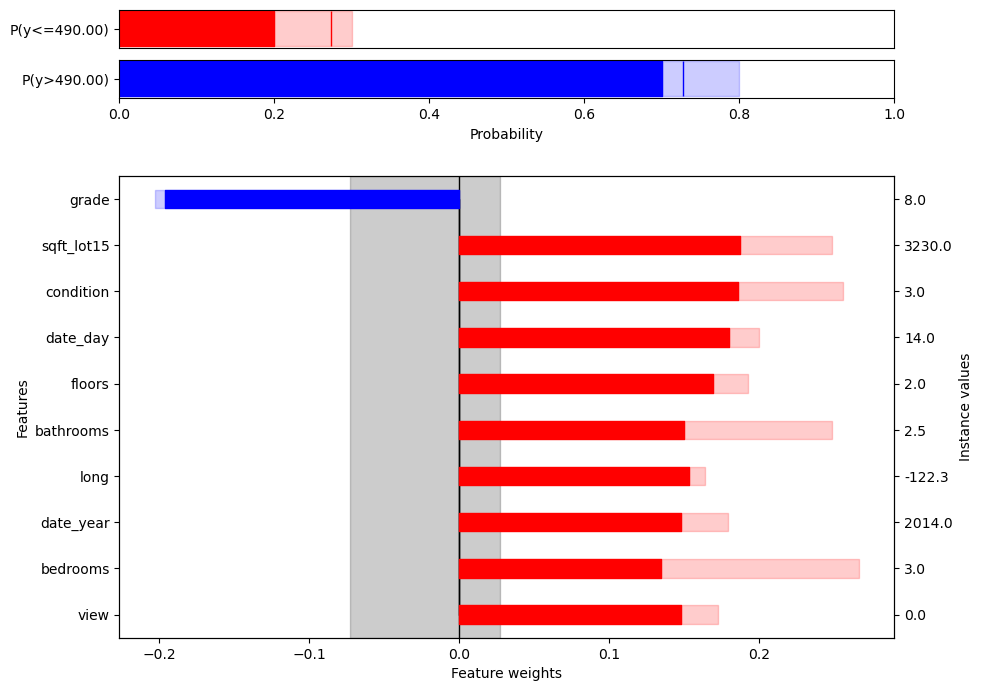}
    \caption{A fast explanation for an instance from the Housing data set, with the threshold $490$}
    \label{fig:housing600}
\end{figure}

The final two examples are taken from the \textit{California Housing} data set with the threshold set to $490$ (which is the median house price in the data set). The example in Figure~\ref{fig:housing600} is likely to be above the threshold with a fairly low uncertainty. One of the features, the \textit{grade}, is favouring the prediction in the sense that if this feature would have been randomly assigned (as it is in the calibration set), it would have resulted in reduced probability $\mathcal{P}(y>490)$. Similarly, there are several features that indicates the opposite impact, even if several of them indicate a lot of uncertainty. It is, however, important to remember that the feature weights are not cumulative, their impact cannot be stacked upon each other. Instead, each feature must be considered by itself. All in all, this is a difficult explanation to interpret, as there are several features that indicate that they favour a lower price. At the same time, the initial probability $\mathcal{P}(y>490)$ is still rather high, reducing $\mathcal{P}(y>490)$ with up to $20$ percentage points (the highest positive weights), would still favour a higher price. In this case, the house was sold for $\$600K$.

\begin{figure}
    \centering
    \includegraphics[width=0.9\linewidth]{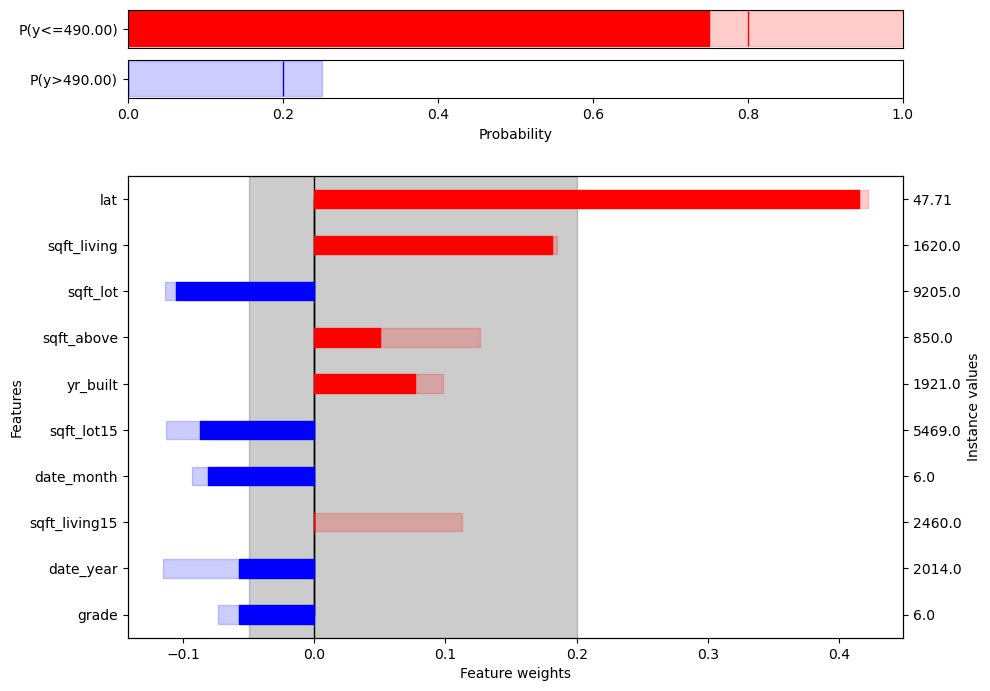}
    \caption{A fast explanation for another instance from the Housing data set, with the threshold $490$}
    \label{fig:housing350}
\end{figure}

The final example, seen in Figure~\ref{fig:housing350}, is also from the same data set, using the same threshold. Here the probability is high for a lower price, even though the uncertainty is also high. One of the feature weights is considerably larger than the other, indicating that the location along the \textit{latitude} is an important feature for the prediction of this instance, strongly favouring the likelihood for a low price. None of the other feature weights have nearly the same impact. In this case, the house was sold for $\$350K$, which seem reasonable given the indication of the model and the explanation. 

These examples have served to indicate how fast explanations can be used to indicate which features that are important. However, it is worth noting that this is all they can do, they cannot provide the user with any additional insights on why or in what way a feature is important. Compared with Calibrated Explanations, having the possibility to answer the why and how questions with their slower explanations, Fast Calibrated Explanations have the benefit of being very fast at the cost of being less informative.  

\section{Concluding Discussion} \label{sec:conclusion}
This paper address a common drawback among explanation methods, namely the computational overhead of explaining an instance. The proposed solution combines fundamental building blocks from two recently proposed explanation methods: Calibrated Explanations and ConformaSight. The proposed solution have used the perturbation strategy used in ConformaSight in combination with the explanation engine in Calibrated Explanations, allowing really fast local explanations with uncertainty quantification for both classification and thresholded regression. The uncertainty quantification extend both the calibrated predictions and the provided feature weights. Fast Calibrated Explanations is able to take advantage of Calibrated Explanations support for both binary and multi-class classification, as well as thresholded regression, providing the probability of the true target being above a user-given threshold. 

Possible directions for future work include considering the issue of perturbations outside the natural scope of the data set as well as ways of speeding up Calibrated Explanations using insights from Fast Calibrated Explanations. Another important area for future work is to consider the decision-making aspect, focusing on situations where fast explanations are critical, exploring how our solution can help create trustworthy explanations in such use cases. Currently, Fast Calibrated Explanations does not convey much insights for standard regression, which should be addressed in future development.

\section*{Acknowledgement}
The Tuwe Löfström and Helena Löfström acknowledge the Swedish Knowledge Foundation and industrial partners for financially supporting the research and education environment on Knowledge Intensive Product Realisation SPARK at Jönköping University, Sweden. Projects: PREMACOP grant no. 20220187, AFAIR grant no. 20200223, and ETIAI grant no. 20230040. The authors Fatima Rabia Yapicioglu and Alessandra Stramiglio are PhD students at the University of Bologna, DISI, funded by PNRR (Piano Nazionale di Ripresa e Resilienza), Italy.

\section*{Declaration of generative AI and AI-assisted technologies in the writing process}
During the preparation of this work the author(s) used ChatGPT and Grammarly in order to improve language and style. After using these tools/services, the author(s) reviewed and edited the content as needed and take(s) full responsibility for the content of the published article.

\bibliographystyle{elsarticle-num-names}

\bibliography{main}
% \begin{thebibliography}{00}

% %% For authoryear reference style
% %% \bibitem[Author(year)]{label}
% %% Text of bibliographic item

% \bibitem[Lamport(1994)]{lamport94}
%   Leslie Lamport,
%   \textit{\LaTeX: a document preparation system},
%   Addison Wesley, Massachusetts,
%   2nd edition,
%   1994.

% \end{thebibliography}
\end{document}